\documentclass[runningheads]{llncs}

\usepackage{graphicx}
\usepackage{amsmath}
\usepackage{amssymb}
\usepackage{algorithm}
\usepackage[noend]{algorithmic}

\usepackage{verbatim}

\usepackage{listings}
\usepackage{xcolor}
\PassOptionsToPackage{hyphens}{url}\usepackage{hyperref}

\definecolor{codegreen}{rgb}{0,0.6,0}
\definecolor{codegray}{rgb}{0.5,0.5,0.5}
\definecolor{codepurple}{rgb}{0.58,0,0.82}
\definecolor{backcolour}{rgb}{0.97,0.97,0.97}

\lstdefinestyle{python_jay}{
    backgroundcolor=\color{backcolour},
    commentstyle=\color{codegreen},
    keywordstyle=\color{blue},
    numberstyle=\tiny\color{codegray},
    stringstyle=\color{codepurple},
    basicstyle=\ttfamily\footnotesize,
    breakatwhitespace=false,
    breaklines=true,
    captionpos=b,
    keepspaces=true,
    numbers=left,
    numbersep=5pt,
    showspaces=false,
    showstringspaces=false,
    showtabs=false,
    tabsize=2
}

\begin{document}

\title{A tale of two toolkits, report the third: on the usage and performance of HIVE-COTE v1.0}

\titlerunning{HIVE-COTE v1.0}

\author{Anthony Bagnall  \and Michael Flynn \and James Large \and Jason Lines \and Matthew Middlehurst}

\authorrunning{Bagnall et al.}

\institute{
Institute\\ University of East Anglia
\email{ajb@uea.ac.uk}
}

\maketitle

\begin{abstract}
The Hierarchical Vote Collective of Transformation-based Ensembles (HIVE-COTE) is a heterogeneous meta ensemble for time series classification. Since it was first proposed in 2016, the algorithm has undergone some minor changes and there is now a configurable, scalable and easy to use version available in two open source repositories. We present an overview of the latest stable HIVE-COTE, version 1.0, and describe how it differs to the original. We provide a walkthrough guide of how to use the classifier, and conduct extensive experimental evaluation of its predictive performance and resource usage. We compare the performance of HIVE-COTE to three recently proposed algorithms.


\end{abstract}

\section{Introduction}
\label{sec:intro}

The Hierarchical Vote Collective of Transformation-based Ensembles (HIVE-COTE) is a heterogeneous meta ensemble for time series classification.
The key principle behind HIVE-COTE is that time series classification (TSC) problems are best approached by careful consideration of the data representation, and that with no expert knowledge to the contrary, the most accurate algorithm design is to ensemble classifiers built on different representations.

HIVE-COTE was first described in an ICDM paper~\cite{lines16hive} which was later expanded to become an ACM Transactions paper~\cite{lines18hive}. At the time, HIVE-COTE was significantly more accurate on average than other known approaches~\cite{bagnall17bakeoff} on the 85 datasets that were then the complete UCR archive~\cite{dau19ucr}.
HIVE-COTE was an improvement over the 2015 version, called just COTE on publication~\cite{bagnall15cote} but later renamed Flat-COTE to differentiate it from its successor. Flat-COTE is a standard ensemble of a range of classifiers built on different representations. It was itself a natural extension of the Elastic Ensemble~\cite{lines15elastic} which only contains nearest neighbour classifiers using different distance measures. HIVE-COTE takes a more structured approach than Flat-COTE. The original HIVE-COTE, which we will henceforth refer to as HIVE-COTE alpha, contained the following classification modules: Elastic Ensemble (EE)~\cite{lines15elastic}; Shapelet Transform Classifier (STC)~\cite{hills14shapelet}; Time Series Forest (TSF)~\cite{deng13forest}; and Bag of Symbolic-Fourier Approximation Symbols (BOSS)~\cite{schafer15boss}. Each module is encapsulated and built  on the train data independently of the others. For new data, each module passes an estimate of class probabilities to the control unit, which combines them to form a single prediction. It does this by weighting the probabilities of each module by an estimate of its testing accuracy formed from the training data.

Our goal with HIVE-COTE alpha was to achieve the highest level of accuracy without concern for the computational resources. This has lead to the perception that HIVE-COTE is very slow and does not scale well. Whilst this is true if used in its basic form, it is in fact very simple to restructure HIVE-COTE so it achieves the same level of accuracy in orders of magnitude less time. We have made many small changes to HIVE-COTE with the goal of making it scalable and more useful. We describe these improvements and encapsulate them as HIVE-COTE version 1.0. The changes in HIVE-COTE are both algorithmic and engineering in nature.

The two slowest components of HIVE-COTE alpha are STC (in its old format) and EE. STC used to conduct a full enumeration of all possible shapelets. We have found that this enormous computational effort is not only unnecessary, but often results in over fitting. EE requires cross validation of numerous nearest neighbour classifiers and is very slow on training and testing. EE resulted from of a comparative study of nearest neighbour distance measures. Our hypothesis was there was no significant difference between the numerous distance measures and dynamic time warping when used in nearest neighbour classifiers, which is true. We only ensembled as an afterthought. We were surprised to see significant improvement. Its design was necessarily ad hoc to avoid over fitting. We have found that dropping EE all together does not make HIVE-COTE much worse. The main changes are:
\begin{enumerate}
\item STC no longer fully enumerates the shapelet space.
    \item EE is dropped altogether from HIVE-COTE.
    \item The STC, BOSS and RISE components include revisions to improve efficiency.
    \item All components and HIVE-COTE 1.0 itself are now contractable (you can set a run time limit), checkpointable (you can save a version to continue building later) and tuneable (select parameters based on train set performance).
    \item HIVE-COTE 1.0 can be threaded, built from existing results and easily configured.
\end{enumerate}

The aim of this report is to describe the changes, showcase the usage of HIVE-COTE and to present some new benchmark results that should be used in all future experiments, and to demonstrate the scalability of HIVE-COTE.

\section{HIVE-COTE 1.0 Design}
\label{sec:design}

\begin{figure}[h]
	\centering
    \includegraphics[width=\linewidth,trim={1cm 2cm 1cm 2cm},clip]{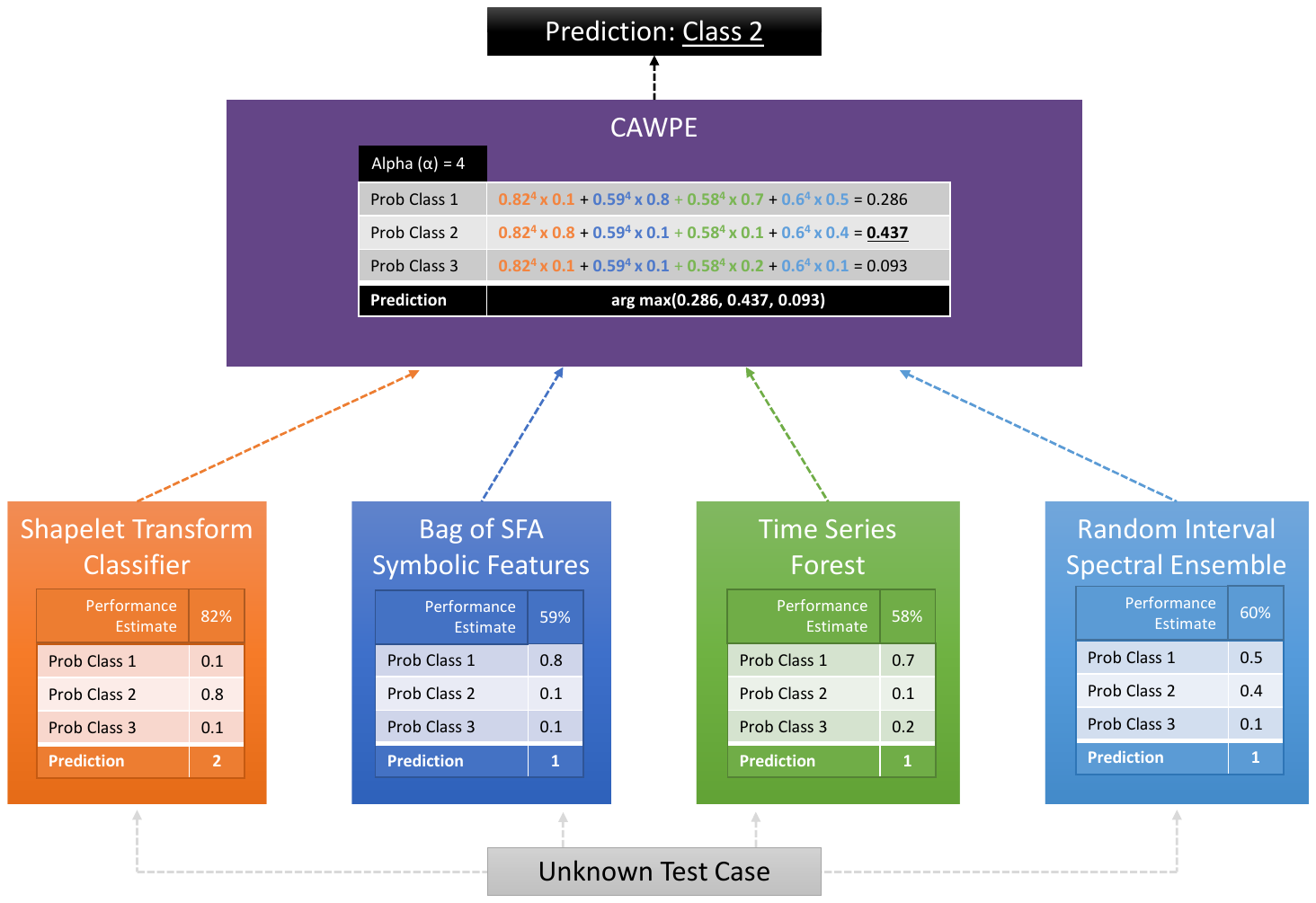}
    \caption{An overview of the ensemble structure of HIVE-COTE 1.0. Each module produces an estimate of the probability of membership of each class. The control unit (CAWPE) combines these probabilities, weighted by an estimate of the quality of the module found on the train data.}
    \label{fig:hive-cote}
\end{figure}

Figure~\ref{fig:hive-cote} provides an overview of the HIVE-COTE structure. The top level ensemble structure and the implementation of each component are described below.
\subsection{Ensemble Structure}

HIVE-COTE adopts the Cross-validation Accuracy Weighted Probabilistic Ensemble (CAWPE) ensemble structure~\cite{large19cawpe}, summmarised in Algorithm~\ref{algo:cawpe}.
\begin{algorithm}[!ht]
 	\caption{HIVE-COTE classify(A test case $\bf{x}$)}
	\label{algo:cawpe}
	\begin{algorithmic}[1]
        \ENSURE prediction for case $\bf{x}$	
        \REQUIRE A set of classifiers $<M_1,\ldots, M_k>$, an exponent $\alpha$, a set of weights $w_i$, and the number of classes $c$
        \STATE $\{\hat{p}_1,\ldots,\hat{p}_c\}= \{0,\ldots,0\}$
	    \FOR {$i\leftarrow 1$ to $k$}
		    \FOR {$j\leftarrow 1$ to $c$}
		        \STATE $\hat{{q_j}} \leftarrow \hat{p}((y=j|M_i,\bf{x})$
    	        \STATE $\hat{p}_j \leftarrow \hat{p}_j+ w_i^\alpha \cdot \hat{q}_{j}$
        	\ENDFOR
        \ENDFOR
        \RETURN $argmax_{j=1 \ldots c}    \hat{p}_j$
 \end{algorithmic}
 \end{algorithm}
CAWPE uses an estimate of the accuracy of each classifier to weight the probability estimates of each component. It constructs a tilted distribution through exponentiation (using $\alpha$) to extenuate differences in classifiers. The weight for each component is found either through ten fold cross validation, or, if the classifier has the capability to estimate its own performance, internally.


\subsection{Time Series Forest (TSF)}
    TSF~\cite{deng13forest} aims to capture basic summary features from intervals of a time series. For any given time series of length $m$ there are $m(m-1)/2$ possible intervals that can be extracted.
    TSF takes a random forest-like approach to sampling these intervals.
        A formal description of TSF is provided in Algorithm~\ref{algo:tsf}.

    For each tree, $r$ intervals are randomly selected (lines 5-7), each with a random start position and length.  Each interval is summarised by the mean, standard deviation and slope (lines 8-11), and the summaries of each interval are concatenated into a single feature vector of length $3r$ for each time series. A decision tree is built on this concatenated feature vector (line 12).
    New cases are classified using a majority vote of all trees in the forest. The version of TSF used in the bake off~\cite{bagnall17bakeoff} employed the random tree used by random forest. However, the decision tree described in~\cite{deng13forest} has some minor differences to the random tree. It makes no difference in terms of accuracy, but the tree from~\cite{deng13forest}, the time series tree, has advantages in terms of interpretability. Hence, the current version of TSF uses the original version, which we call TimeSeriesTree.
Contracting is enforced by the method timeRemaining. TSF simply builds until it has the required number of trees or the time runs out.



    \begin{algorithm}[htb]
        \caption{buildTSF(A list of $n$ cases length $m$, ${\bf T}=({\bf X,y})$)}
        \label{algo:tsf}
        \begin{algorithmic}[1]
            \REQUIRE the number of trees, $k$; the minimum interval length, $p$;  the number of intervals per tree, $r$. (default $k \leftarrow 500$, $p\leftarrow 3$, and $r \leftarrow \sqrt{m}$)
            \STATE Let ${\bf F} = ({\bf F_1} \ldots {\bf F_k})$ be the trees in the forest
        	\STATE $ i \leftarrow 1 $
        	\WHILE {$i < k$ $\AND$ timeRemaining()}
                \STATE Let ${\bf S}$ be a list of $n$ cases $({\bf s_1} \ldots {\bf s_n})$ with $3r$ attributes
        		\FOR {$j \leftarrow 1$ to $r$}
        		    \STATE $b \leftarrow $ randBetween$(1,m-p)$
        		    \STATE $e \leftarrow $ randBetween$(b+p,m)$
        		    \FOR {$t \leftarrow 1$ to $n$}
        		        \STATE $s_{t,3(j-1)+1} \leftarrow$ mean$({\bf x}_t,b,e)$
        		        \STATE $s_{t,3(j-1)+2} \leftarrow$ standardDeviation$(\bf{x}_t,b,e)$
        		        \STATE $s_{t,3(j-1)+3} \leftarrow$ slope$({\bf x}_t,b,e)$
        		    \ENDFOR
        		\ENDFOR
        		\STATE $\bf{F_i}.$buildTimeSeriesTree(${\bf S},y$)
            \ENDWHILE
        \end{algorithmic}
    \end{algorithm}

\subsection{Random Interval Spectral Ensemble (RISE)}
Like TSF, RISE~\cite{lines18hive} is a tree based interval ensemble. Unlike TSF, it uses a single interval for each tree, and it uses spectral features rather than summary statistics.  RISE was recently updated to be faster and contractable~\cite{flynn19contract}. During the build process, summarised in Algorithm \ref{algo:rise_build}, a single random interval is selected for each tree. The first tree is a special case in which the whole series is used (lines 5 and 6). Otherwise, an interval with length that is a power of 2 (line 9) is chosen. The same interval for each series is then transformed using the Fast Fourier Transform (FFT) and Auto Correlation Function (ACF). This is a change on the original RISE which also used the partial autocorrelation function (PACF) and autoregressive (AR) model features. Restriction to the Fourier and ACF coefficients does not decrease accuracy, but makes the algorithm much faster. The power spectrum coefficients are concatenated with the first 100 ACF coefficients to form a new training set. In the \texttt{tsml} implementation of RISE the base classifier used is the RandomTree classifier used by random forest (line 13). In the test process class probabilities are assigned as a proportion of base classifier votes.

RISE controls the contract run time by creating an adaptive model of the time to build a single tree (lines 4 and 14). This is important for long series (such as audio), where very large intervals can mean very few trees. Details are in~\cite{flynn19contract}.

\begin{algorithm}
	\caption{buildRISE(A list of $n$ cases of length $m$, ${\bf T}=({\bf X,y})$)}
    \label{algo:rise_build}
	\begin{algorithmic}[1]
	\REQUIRE the number of trees, $k$; the minimum interval length, \textit{p}. (default $k \leftarrow 500$, $p\leftarrow $min($16,m/2$))
	\STATE Let ${\bf F} \leftarrow<{\bf F_1} \ldots {\bf F_{k}}>$ be the trees in the forest.
	\STATE $ i \leftarrow 1 $
	\WHILE {$i < k$ $\AND$ timeRemaining()}
        \STATE buildAdaptiveTimingModel()
        \IF{$i = 1$}
            \STATE $r \leftarrow m$
        \ELSE
        	\STATE $max \leftarrow$ findMaxIntervalLength()
            \STATE $r \leftarrow$ findPowerOf2Interval($p$, $max$)
		\ENDIF
		\STATE $b \leftarrow$ randBetween($1,m-r$)
		\STATE $\bf{T'}  \leftarrow$ removeAttributesOutsideOfRange($\bf{T}, b, r$)
    	\STATE $\bf{S}  \leftarrow$ getSpectralFeatures($\bf{T'}$)
		\STATE ${\bf F_i}$.buildRandomTreeClassifier($\bf{S}$,$y$)
        \STATE updateAdaptiveModel($r$)
        \STATE $i \leftarrow i +1 $
	\ENDWHILE
	\end{algorithmic}
\end{algorithm}

\subsection{Bag of SFA Symbolic Features (BOSS)}

Dictionary based classifiers convert real valued time series into a sequence of discrete symbol words, then base classification on these words. Commonly, a sliding window of length $w$ is run across a series. For each window, the real valued series of length $w$ is converted through approximation and discretisation processes into a symbolic string of length $l$ (referred to as a word), which consists of $\alpha$ possible letters. The occurrence in a series of each word from the dictionary defined by $l$ and $\alpha$ is counted and, once the sliding window has completed, the series is transformed into a histogram. Classification is based on the histograms of the words extracted from the series, rather than the raw data.  The Bag of Symbolic-Fourier-Approximation Symbols (BOSS)~\cite{schafer15boss} ensemble was found to be the most accurate dictionary based classifier in the bake off~\cite{bagnall17bakeoff}. Hence, it forms our benchmark for new dictionary based approaches.

Algorithm~\ref{alg:boss} gives a formal description of the bag forming process of an individual BOSS classifier. Windows may or may not be normalised (lines 6 and 7). Words are created using Symbolic Fourier Approximation (SFA)~\cite{schafer12sfa} (lines 8-13). SFA first finds the Fourier transform of the window (line 8), ignoring the first term if normalisation occurs (lines 9-12). It then discretises the first $l$ Fourier terms into $\alpha$ symbols to form a word in the method SFAlookup, using a bespoke supervised discretisation algorithm called Multiple Coefficient Binning (MCB) (line 13). Lines 14-16 implement the process of not counting self similar words: if two consecutive windows produce the same word, the second occurrence is ignored. This is to avoid a slow-changing pattern relative to the window size being over-represented in the resulting histogram.

BOSS uses a non-symmetric distance function in conjunction with a nearest neighbour classifier. Only the words contained in the test instance's histogram (i.e. the word count is above zero) are used in the distance calculation, but it is otherwise the Euclidean distance.

    \begin{algorithm}[ht]
    	\caption{baseBOSS(A list of $n$ time series of length $m$, ${\bf T}=({\bf X,y})$)}
    	\label{alg:boss}
    	\begin{algorithmic}[1]
    \REQUIRE the word length $l$, the alphabet size $\alpha$, the window length $w$, normalisation parameter $z$
    		\STATE Let ${\bf H}$ be a list of $n$ histograms $({\bf h}_1,\ldots,{\bf h}_n)$
    		\STATE Let ${\bf B}$ be a matrix of $l$ by $\alpha$ breakpoints found by MCB
    		\FOR {$i \leftarrow  1$ to $n$}
    			\FOR {$j \leftarrow 1$ to $m-w+1$}
    				\STATE ${\bf s}\leftarrow x_{i,j} \ldots x_{i,j+w-1}$
    				\IF{$z$}
    				    \STATE $s \leftarrow $normalise({\bf $s$})
    				\ENDIF
    				\STATE ${\bf q} \leftarrow$ DFT(${\bf s}, l, \alpha$,$p$) \COMMENT{ {\em {\bf q} is a vector of the complex DFT coefficients}}
    				\IF{$z$}
        			    \STATE ${\bf q'} \leftarrow (q_2 \ldots q_{l/2+1})$
        			\ELSE
        			    \STATE ${\bf q'} \leftarrow (q_1 \ldots q_{l/2})$
        			\ENDIF
    				\STATE ${\bf r} \leftarrow$ SFAlookup(${\bf q', B}$)
    				\IF{${\bf r} \neq {\bf p}$}
    					\STATE $pos \leftarrow $index(${\bf r}$)
    					\STATE ${h}_{i,pos} \leftarrow {h}_{i,pos} + 1$
    				\ENDIF
    				\STATE ${\bf p} \leftarrow {\bf r} $
    			\ENDFOR
    		\ENDFOR
    	\end{algorithmic}
    \end{algorithm}

The final classifier is an ensemble of individual BOSS classifiers. The original BOSS ensemble built all models over a pre-defined parameter space for $w$, $l$, $z$ and $\alpha$ and retained all base classifiers with accuracy of 92\% or higher of the best. This introduces instability in memory usage and carries a time overhead.
HIVE-COTE 1.0 uses contractable BOSS (cBOSS)~\cite{middlehurst19scalable} as its dictionary based classifier. cBOSS changes the method used by BOSS to form its ensemble to improve efficiency and allow for a number of usability improvements. cBOSS was shown to be an order of magnitude faster than BOSS on both small and large datasets from the UCR archive while showing no significant difference in accuracy~\cite{middlehurst19scalable}. It randomly samples the parameter space without replacement (line 8), subsamples the data for each base classifier (line 10), and retains a fixed number of base classifiers. An exponential weighting scheme based on train accuracy, such as the one used in HIVE-COTE, is introduced for ensemble members.

\begin{algorithm}[ht]
\caption{cBOSS(A list of $n$ cases length $m$, ${\bf T}=({\bf X,y})$)}
\label{alg:cBOSS}
\begin{algorithmic}[1]
\REQUIRE the maximum number of base classifiers, $k$; the number of parameter samples, $s$; the proportion of cases to sample, $p$.
        (default $k \leftarrow 50$; $s \leftarrow 250$; $p \leftarrow 0.7$)
\STATE Let $w$ be window length, $l$ be word length, $z$ be normalise/not normalise and $\alpha$ be  alphabet size.
\STATE Let ${\bf B} \leftarrow<{\bf B_1} \ldots {\bf B_{k}}>$ be a list of $k$ BOSS classifiers
\STATE Let ${\bf W} \leftarrow <{w}_1,\ldots,{w}_k>$ be a list of classifier weights
\STATE Let ${\bf R}$ be a set of possible BOSS parameter combinations
\STATE $i \leftarrow 0$
\STATE $minAcc \leftarrow \infty$
\WHILE {$i < s\; \AND\; |{\bf R}| > 0 \; \AND\;$ timeRemaining()}
    \STATE $[l,a,w,p] \leftarrow$ randomSampleParameters(${\bf R}$)
    \STATE $ {\bf R} = {\bf R} \setminus\{[l,a,w,p]\} $
    \STATE ${\bf T'} \leftarrow$ subsampleData(${\bf T}$)
    \STATE ${\bf cls} \leftarrow$ baseBOSS(${\bf T'},l,a,w,z$)
    \STATE $acc \leftarrow$ estimateAccuracy(${\bf T'}$,{\bf $cls$}) \COMMENT{ {\em estimate accuracy on train data}}
    \IF{$i < k$}
        \STATE ${\bf B_i} \leftarrow cls$, $w_i \leftarrow acc^4$
        \IF{$acc < minAcc$}
            \STATE $minAcc \leftarrow acc$, $idx \leftarrow i$
        \ENDIF
    \ELSIF{$acc > min\_acc$}
        \STATE ${\bf B_{idx}} \leftarrow {\bf cls}$, $w_{idx} \leftarrow acc^4$
        \STATE $[minAcc,idx] \leftarrow$ findLowestAcc(${\bf B}$)
        		    \ENDIF
        		    \STATE $i \leftarrow i+1$
        		\ENDWHILE
        	\end{algorithmic}
        \end{algorithm}

\subsection{Shapelet Transform Classifier (STC)}

There are two significant changes to the STC used in HC 1.0. Firstly, it only fully enumerates the shapelet space when there is sufficient time to do so. Secondly, it uses a Rotation Forest classifier~\cite{rodriguez06rotf} rather than a heterogeneous ensemble. The shapelet transform is highly configurable: it can use a range of sampling/search techniques in addition to alternative quality measures. We present the default settings and direct the interested reader to the code. The amount of time for the shapelet search is now a parameter. The algorithm calculates how many possible shapelets there are in a data set, then estimates how many shapelets it can sample from each series. After searching, it updates its timing model using simple reinforcement learning. These operations are encapsulated in operations estimateNumberOfShapelets (line 7) and  updateTimingModel (line 11). Shapelets are randomly sampled in the method sampleShapelets (line 8). If the algorithm is allowed more shapelets than the series contains, it evaluates them all. We have experimented with a range of alternative neighbourhood search algorithms, but nothing is much better than random search. Once the shapelets are generated, they are evaluated using information gain (line 9). We use a one vs all evaluation for multi-class problems~\cite{bostrom17binary}. Overlapping shapelets are removed in line 10, before the candidates are merged into the overall pool, with the weakest members of the population being deleted. Once the search is complete, the transform is performed (line 13) and the classifier constructed (line 14).

        \begin{algorithm}[!htb]
        	\caption{STC(A list of $n$ cases length $m$, ${\bf T}=({\bf X,y})$)}
        	\label{alg:STC}
        	\begin{algorithmic}[1]
        \REQUIRE the maximum number of shapelets to keep, $k$; the shapelet search time, $t$. (default $k \leftarrow 1000$, $t \leftarrow 1$ hour.
        \STATE Let ${\bf S}$ be a list of up to $k$ shapelets
        \STATE Let ${\bf R}$ be a rotation forest classifier.
        \STATE $i \leftarrow 0$
        \STATE $minIG \leftarrow 0$
        \STATE
        \WHILE {shapeletTimeRemaining($t$)}
            \STATE $p \leftarrow$ estimateNumberOfShapelets($t, m, n$)
            \STATE ${\bf S'} \leftarrow$ sampleShapelets(${\bf x_i}$,$p$)
            \STATE ${\bf s} \leftarrow$ evaluateShapelets($S'$,${\bf T}$)
            \STATE ${\bf S'} \leftarrow$ removeSelfSimilar({\bf$S'$}, $s$)
            \STATE updateTimingModel()
            \STATE ${\bf S} \leftarrow $ merge({\bf $S$},{\bf $S'$})
        \ENDWHILE
        \STATE ${\bf X'} \leftarrow $ shapeletTransform({\bf$X$},{\bf$S$})
        \STATE ${\bf R}$.buildRotationForest(${\bf X'},${\bf y})
        	\end{algorithmic}
        \end{algorithm}

\section{HIVE-COTE 1.0 Usability}

We help maintain two toolkits that include time series classification functionality. \texttt{aeon}\footnote{\url{https://github.com/aeon-toolkit/aeon}} is an open source, Python based, sklearn compatible toolkit for time series analysis. \texttt{aeon} is designed to provide a unifying API for a range of time series tasks such as annotation, prediction and forecasting (see~~\cite{bagnall19toolkits} for an experimental comparison of some of the classification algorithms available). The Java toolkit for time series machine learning, \texttt{tsml}\footnote{\url{https://github.com/uea-machine-learning/tsml}}, is Weka compatible and is the descendent of the codebase used to perform the bake off. The two toolkits will eventually converge to include all the features described here.  Experiments reported in this paper are conducted with \texttt{tsml}, as it has more functionality.

\subsection{Java Implementation of HIVE-COTE 1.0 in \texttt{tsml}}

The \texttt{HIVE\_COTE} class is in the package \texttt{tsml.classifiers.hybrids} and can be used as any other Weka classifier. The default configuration is that described in this paper.
The code described here is all available in the class \texttt{EX07\_HIVE\_COTE\_Examples} with more detail and comments. A basic build is described in Listing~\ref{hive_java}. It cannot handle missing values, unequal length series or multivariate problems yet.
\lstset{style=python_jay}
\begin{lstlisting}[language=Java, caption=A most basic way using HIVE-COTE 1.0 in \texttt{tsml}, label=hive_java]
    HIVE_COTE hc = new HIVE_COTE();
    //this setup called in default constructor in April 2020
    hc.setupHIVE_COTE_1_0();
    Instances[] trainTest =
       DatasetLoading.sampleItalyPowerDemand(0);
    hc.buildClassifier(trainTest[0]);
    int correct=0;
    for(Instance ins: trainTest[1]){
        double c=hc.classifyInstance(ins);
        if(c==ins.classValue())
            correct++;
\end{lstlisting}
We rarely build the classifier in this way. Instead, we build the components and post process the meta ensemble. This is most easily done using our \texttt{Experiments} class, which formats the output in a standard way. An example code snippet is in Listing~\ref{hive_experiments}. Details on optional input flags not given below can be found in the code.
\begin{lstlisting}[language=Java, caption=Using Experiments.java to build a single component., label=hive_experiments]
    String[] settings=new String[6];
    //Where to get data
    settings[0]="-dp=src/main/java/experiments/data/tsc/";
    //Where to write results
    settings[1]="-rp=Temp/";
    //Whether to generate train files or not
    settings[2]="-gtf=true";
    //Classifier name: See ClassifierLists for valid options
    settings[3]="-cn=TSF";
    //Problem file
    settings[4]="-dn=Chinatown";
    //Resample number: 1 gives the default train/test split
    settings[5]="-f=1";
    Experiments.ExperimentalArguments expSettings =
        new Experiments.ExperimentalArguments(settings);
    Experiments.setupAndRunExperiment(expSettings);
\end{lstlisting}
\texttt{HIVE\_COTE} can read component results directly from file using syntax of the form given in Listing~\ref{hive_from_file}. It will look in the directory structure created by Experiments. In this example, when building TSF, it will look for \\{\em ``C:/Temp/TSF/Predictions/Chinatown/trainFold0.csv"}. Currently, this file loading method requires all the classifier results to be present in order to build.
\begin{lstlisting}[language=Java, caption=Buildind HIVE-COTE from existing results files, label=hive_from_file]
    HIVE_COTE hc=new HIVE_COTE();
    hc.setBuildIndividualsFromResultsFiles(true);
    hc.setResultsFileLocationParameters("C:/Temp/", "Chinatown", 0);
    String[] components={"TSF","RISE","cBOSS","STC"};
    hc.setClassifiersNamesForFileRead(components);
\end{lstlisting}
\texttt{HIVE\_COTE} is configurable for different components, threadable (see Listing~\ref{hive_configure}) and contractable (see Listing~\ref{hive_contract}). In sequential mode, it simply divides the time equally between components. When threaded, it gives the full contract time to each component.   It does not yet thread individual components; it is on our development list.
You can set the maximum build time for \texttt{HIVE\_COTE}  if the components all implement the \texttt{TrainTimeContractable} interface.

\begin{lstlisting}[language=Java, caption=Threaded build of HIVE-COTE with bespoke classifiers, label=hive_configure]
    HIVE_COTE hc = new HIVE_COTE();
    EnhancedAbstractClassifier[] c=new EnhancedAbstractClassifier[2];
    c[0]=new RISE();
    c[1]=new TSF();
    String[] names={"RISE","TSF"};
    hc.setClassifiers(c,names,null);
    hc.enableMultiThreading(2);
\end{lstlisting}
\begin{lstlisting}[language=Java, caption=Contracting HIVE-COTE for a rom existing results files, label=hive_contract]
//Ways of setting the contract time
    HIVE_COTE hc = new HIVE_COTE();
//Minute, hour or day limit
    hc.setMinuteLimit(1);
    hc.setHourLimit(2);
    hc.setDayLimit(1);
//Specify units
    hc.setTrainTimeLimit(30, TimeUnit.SECONDS);
    hc.setTrainTimeLimit(1, TimeUnit.MINUTES);
//Or just give it in nanoseconds
    hc.setTrainTimeLimit(10000000000L);
    hc.buildClassifier(train);
\end{lstlisting}
Finally, \texttt{HIVE\_COTE} is tuneable. Our method of implementing tuned classifiers is to wrap the base classifier in a \texttt{TunedClassifier} object which interacts through the method \texttt{setOptions}. An example of tuning the $\alpha$ parameter is given in Figure~\ref{hive_tuning}. We advise tuning using results loaded from file. We have found tuning $\alpha$ makes no significant difference. We have not finished evaluating tuning which components to use.

\begin{lstlisting}[language=Java, caption=Tuning HIVE-COTE $\alpha$ parameter from existing results files, label=hive_tuning]
    HIVE_COTE hc=new HIVE_COTE();
    hc.setBuildIndividualsFromResultsFiles(true);
    hc.setResultsFileLocationParameters(resultsPath, dataset, fold);
    hc.setClassifiersNamesForFileRead(cls);
    TunedClassifier tuner=new TunedClassifier();
    tuner.setClassifier(hc);
    ParameterSpace pc=new ParameterSpace();
    double[] alphaVals={1,2,3,4,5,6,7,8,9,10};
    pc.addParameter("A",alphaVals);
    tuner.setParameterSpace(pc);
\end{lstlisting}

\subsection{Python Implementation of HIVE-COTE 1.0 in \texttt{aeon}}

A Python implementation of HIVE-COTE is under development and available in \texttt{aeon}. As previously discussed, this version of the algorithm is less mature in terms of its development and the results reported in this paper are from the Java version of the algorithm. The \texttt{aeon} implementation is in an alpha state, and will eventually converge on the same functionally as the more developed Java implementation, but currently has a number of limitations (such as building from file and running constituents in parallel). Further, the Python implementations of the constituent classifiers are less efficient than the Java implementations, and as such, HIVE-COTE 1.0 in \texttt{aeon} is slower than the \texttt{tsml} implementation on equivalent inputs.

The interface and basic usage of HIVE-COTE in \texttt{aeon} is very similar to that of \texttt{tsml}. The terminology is slightly different however as the \texttt{aeon} version uses fit and predict derived from scikit-learn while the Java version uses build and classify from Weka/\texttt{tsml}. Notionally the process of constructing and making predictions with HIVE-COTE are equivalent and a simple example of fitting and predicting with HIVE-COTE in \texttt{aeon} is given in Listing~\ref{hive_python}. 
\lstset{style=python_jay}

\begin{lstlisting}[language=Python, caption=A simple example of using HIVE-COTE 1.0 in \texttt{aeon}, label=hive_python]
def basic_hive_cote(data_dir, dataset_name):
    from aeon.classification.hybrid import HIVECOTEV1
    from aeon.datasets import load_from_tsfile

    # using the default constructor for the HIVE-COTE class
    hc = HIVECOTEV1()

    # loading training data
    train_x, train_y = _(
        data_dir + dataset_name + "_TRAIN.ts")

    # building HIVE-COTE 1.0 sequentially
    hc.fit(train_x, train_y)

    # loading testing data
    test_x, test_y = _(
        data_dir + dataset_name + "_TEST.ts")

    # predict class values of the test data
    preds = hc.predict(train_x)

    # calculate the test accuracy
    acc = accuracy_score(self.train_y, preds)

\end{lstlisting}

\section{Performance}

\begin{figure}[htb]
	\centering
    \includegraphics[width=\linewidth,trim={0cm 8cm 0cm 5cm},clip]{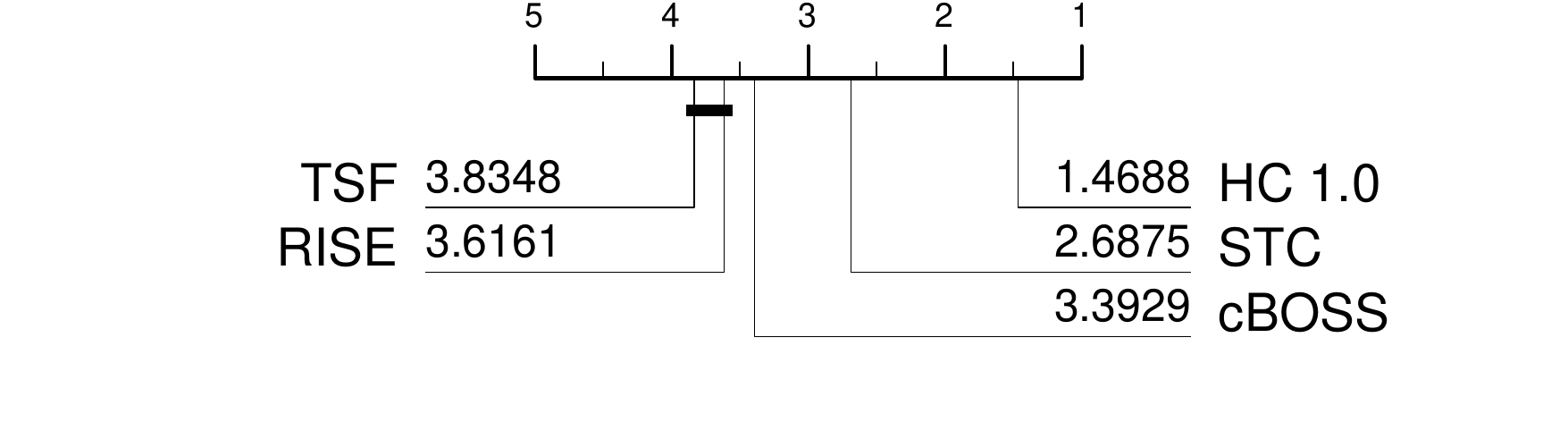}
    \caption{Critical difference diagram for HIVE-COTE 1.0 and its four components on 112 UCR TSC problems. Full results are available from www.timeseriesclassification/Results/HC1.0.xls}
    \label{fig:components}
\end{figure}

To measure performance of the new HIVE-COTE, we evaluate each component and the algorithm itself on 112 of the 128 UCR archive datasets. These 112 datasets are all equal length and have no missing values. Figure~\ref{fig:components} shows the critical difference diagram for HIVE-COTE 1.0 (HC 1.0) and its four components. This broadly mirrors the performance presented in~\cite{lines18hive}.

\begin{figure}[!htb]
	\centering
    \includegraphics[width=\linewidth,trim={0cm 8cm 0cm 5cm},clip]{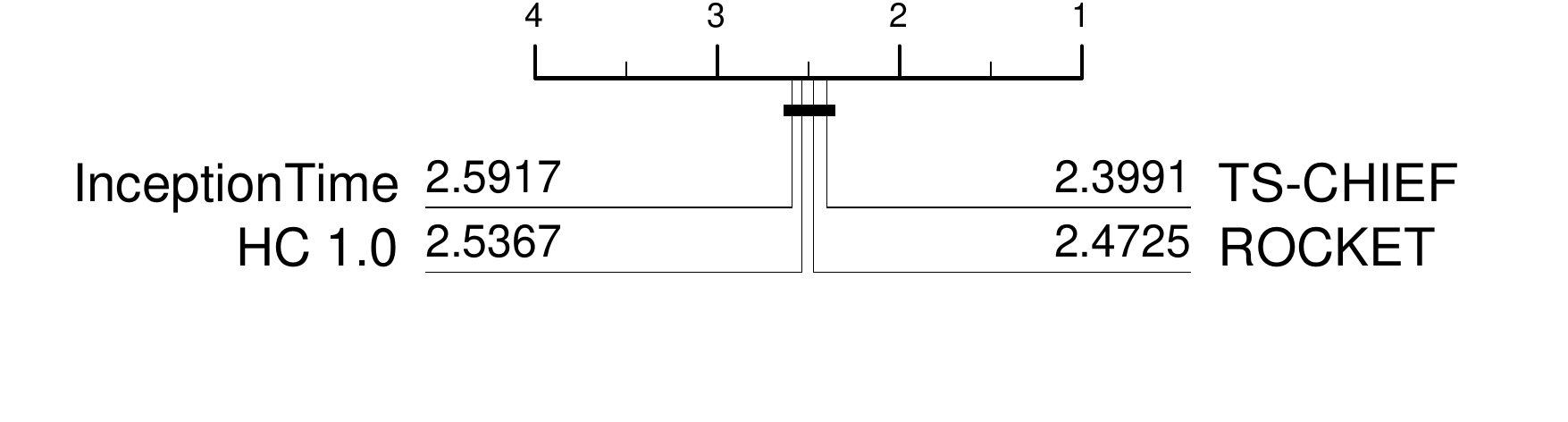}
    \caption{Critical difference diagram for current state of the art on 109 UCR TSC problems.}
    \label{fig:sota}
\end{figure}

We have compared the results of the four components to the original results and found there is no significant different. Comparison of HIVE-COTE alpha version and 1.0 identify a small, but significant, difference. Removing EE makes HIVE-COTE worse on 48 and better on 33 of the 85 datasets used in the original experiments. The mean reduction in accuracy is 0.6\%. The differences in accuracy on specific problems identify those where a distance based approach may be the best. MedicalImages, SonyAIBORobotSurface1, WordSynonyms and Lightning7 were all more than 5\% less accurate with EE removed.  We are trading this small loss in average accuracy for significant gains in run time and reduction in memory usage. We have explored ways of making EE more efficient, but as yet none of these approaches have met the criteria for maintaining accuracy and providing sufficient speed up.
Since HIVE-COTE alpha was proposed, three new algorithms have achieved equivalent accuracy. TS-CHIEF~\cite{shifaz19ts-chief} is a tree ensemble that embeds dictionary, spectral and distance based representations. InceptionTime~\cite{fawaz19inception} is a deep learning ensemble, combining five homogeneous networks each with random weight initialisations for stability. ROCKET~\cite{dempster19rocket} uses a large number (10,000 by default) of randomly parameterised convolution kernels in conjunction with a linear ridge regression classifier. We use the configurations of each classifier described in their respective publications. Figure~\ref{fig:sota} shows the critical difference diagram for these three classifiers and HC 1.0. There is no significant difference between any of them.
The differences between HIVE-COTE and the other three are summarised in Table~\ref{tab:differences}. TS-CHIEF is the most similar to HIVE-COTE, with an average accuracy just 0.25\% lower and a high correlation between accuracies. ROCKET has a high variation in performance, as reflected in the high standard deviation of differences and the difference between the mean and median of the differences. It also has the lowest correlation to HC 1.0.

\begin{table}[!htb]
    \centering
    \caption{Summary of differences in accuracy between HIVE-COTE and the other three algorithms. A negative difference indicates HIVE-COTE is more accurate.}
\begin{tabular}{c|c|c|c|c}
                &   Mean    &  Median       &   Std Dev of differences & Correlation            \\ \hline
TS-CHIEF        & -0.25\%	& 0.00\%        &   	3.801           & 95.54\%   \\
InceptionTime   & -0.65\%   & 0.00\%	    &       5.82	        & 89.46\%\\
ROCKET          & -1.41\%	& 0.05\%	    &       8.64	        & 80.28\%
\end{tabular}
    \label{tab:differences}
\end{table}
Table~\ref{tab:rocket} shows the results for problems with the 10 biggest differences between HIVE-COTE and ROCKET.
ROCKET does very poorly on some problems (hence the large average difference in Table~\ref{tab:differences}). InceptionTime also shares this characteristic of occasionally simply failing on a problem. For example, InceptionTime is over 20\% less accurate on the problem Rock than the other three classifiers. It is not obvious why this happens to both ROCKET and InceptionTime. It may be the result of over fitting.
\begin{table}[!htb]
    \centering
    \caption{Accuracy for ten problems with the biggest difference between HIVE-COTE and ROCKET (five negative and five positive differences).}
\begin{tabular}{l|c|c|c|c}
                      &	TS-CHIEF&	ROCKET	&HC 1.0	& InceptionTime\\ \hline
PigAirwayPressure     & 96.01\% & 	19.55\% & 	95.77\% & 	92.21\%\\
SemgHandMovementCh2   & 88.50\% & 	65.26\% & 	88.90\% & 	55.10\%\\
EthanolLevel          & 60.56\% & 	62.53\% & 	84.90\% & 	87.55\%\\
CinCECGTorso          & 95.34\% & 	86.41\% & 	99.37\% & 	83.28\%\\
ScreenType            & 59.42\% & 	60.90\% & 	72.42\% & 	70.56\%\\ \hline
FiftyWords            & 84.27\% & 	82.51\% & 	77.16\% & 	82.68\%\\	
ChlorineConcentration & 66.08\% & 	79.61\% & 	73.39\% & 	86.36\%\\
MedicalImages         & 79.91\% & 	80.51\% & 	74.04\% & 	79.63\%\\
WordSynonyms          & 79.37\% & 	76.44\% & 	69.32\% & 	75.18\%\\
SonyAIBORobotSurface1 & 88.97\% & 	95.81\% & 	82.63\% & 	95.42\%\\
\end{tabular}
    \label{tab:rocket}
\end{table}

It is worthwhile considering the run time of these algorithms. However, comparisons are made more difficult because of the different software. HIVE-COTE and TS-CHIEF are both built using \texttt{tsml}, so are directly comparable. Table~\ref{tab:train_time} summarises the time taken to train the classifiers on 109 UCR problems. Three problems (HandOutlines, NonInvasiveFetalECGThorax1 and NonInvasiveFetalECGThorax2) are omitted because TS-CHIEF did not complete within 7 days (the job limit on our cluster). Of the HC-1.0 components, STC is by far the slowest. This is caused by the classifier, Rotation Forest, not the transform, which is contracted to take at most one hour. The STC design principle is to choose a large number of shapelets (up to 1000) and let the classifier sort out their importance. Rotation forest is the best approach for problems with all real valued attribute~\cite{bagnall18rotf}, and we have developed a contracted version that can, if necessary, be used to control the build time (see Listing~\ref{hive_tuning}). Most of the extra computation required by HIVE-COTE is in forming the estimates of the accuracy on the train data. TS-CHIEF is the slowest algorithm, but it is approximately the same as HIVE-COTE.

\begin{table}[!htb]
    \centering
    \caption{Time in hours to train a classifier for 112 of the UCR problems}
\begin{tabular}{c|c|c|c|c|c|c}
        & TSF       & cBOSS     &   RISE    &   STC    &    HC 1.0 & TS-CHIEF \\ \hline
Mean    & 	0.13	&   0.11    &	0.15	&   2.30    &	4.26   & 5.75 \\ \hline
Total   &  	14.15	&   12.31	&   16.05	&   251.12	&   464.49 & 626.33  \\    \hline
\end{tabular}
    \label{tab:train_time}
\end{table}

Test time can be a factor for deploying classifiers in time critical situations. Table~\ref{tab:test_time} summarises the time (in minutes) taken to predict the test cases. STC is the slowest component when testing, again caused by the classifier not the transform. TS-CHIEF is the slowest in testing.
\begin{table}[!htb]
    \centering
    \caption{Time in minutes to make predictions on the test data for 109 of the UCR problems}
\begin{tabular}{c|c|c|c|c|c|c}
        & TSF       & cBOSS     &   RISE    &   STC    &    HC 1.0 & TS-CHIEF \\ \hline
Mean    & 	0.01	&   0.08    &	0.07	&   0.62    &	0.78   & 4.09 \\ \hline
Total   &  	1.42	&   8.63	&   7.09	&   67.50	&   84.64  & 445.55  \\    \hline
\end{tabular}
    \label{tab:test_time}
\end{table}
We do not have reliable times for ROCKET, which is run in \texttt{aeon}. It is undoubtedly faster than HIVE-COTE and TS-CHIEF though. We can also measure maximum memory usage of the classifiers, as this is also often a serious bottleneck for scalability. Table~\ref{tab:memory} summarises the memory usage of the six \texttt{tsml} classifiers. The pattern is similar to that observed with run time. TSF, RISE and cBOSS have a light memory footprint. STC, and hence HIVE-COTE, have a larger memory usage. TS-CHIEF is the most memory hungry, and it seemingly does not scale well in terms of memory. For, example, on the 11 problems where HIVE-COTE required more than 1 GB, TS-CHIEF required approximately four times the memory. The highest memory usage on the 109 problem it could complete was 18 GB on FordA and FordB. It seems highly likely that the memory usage of TS-CHIEF could be improved without loss of accuracy.
\begin{table}[htb]
    \centering
    \caption{Memory usage in MB for 109 UCR problems}
\begin{tabular}{c|c|c|c|c|c|c}
        & TSF   & cBOSS     & RISE & STC    & HC 1.0 & TS-CHIEF \\ \hline
Mean    & 162	&   255     & 263 & 1,464 & 1,618      & 2,004          \\
Max     & 1061	&   3432	& 740 &	3,533 & 4,426      & 18,532        \\    \hline
\end{tabular}
    \label{tab:memory}
\end{table}

\section{Conclusions}

The purpose of this report is to present a more practical version of HIVE-COTE and compare it to recent advances in the field of TSC. On average, there is no real difference between InceptionTime, TS-CHIEF, ROCKET and HIVE-COTE in terms of accuracy. ROCKET is undoubtedly the fastest, but it is prone to fail badly on the occasional data set. InceptionTime is slow and requires a GPU. It also fails on some data sets. HIVE-COTE and TS-CHIEF are broadly comparable. HIVE-COTE is presently more configurable and controllable.

We hope the results presented here and on the accompanying website serve to act as a baseline comparison for any new research in the field in the future. We have presented HIVE-COTE as is, without improvements that we have been working on. Our next goal is to find a more accurate version that is comparable in run time and memory usage, and to demonstrate scalability by introducing some much larger problems into the UCR archive. We are also developing versions that can handle unequal length series and multivariate TSC problems.

\bibliographystyle{plain}

\begin{thebibliography}{10}

\bibitem{bagnall19toolkits}
A.Bagnall, F.~Kir\'aly, M.~L\"oning, M.~Middlehurst, and G.~Oastler.
\newblock A tale of two toolkits, report the first: benchmarking time series
  classification algorithms for correctness and efficiency.
\newblock {\em ArXiv e-prints}, arXiv:1909.05738, 2019.

\bibitem{bagnall18rotf}
A.~Bagnall, A.~Bostrom, G.~Cawley, M.~Flynn, J.~Large, and J.~Lines.
\newblock Is rotation forest the best classifier for problems with continuous
  features?
\newblock {\em ArXiv e-prints}, arXiv:1809.06705, 2018.

\bibitem{bagnall17bakeoff}
A.~Bagnall, J.~Lines, A.~Bostrom, J.~Large, and E.~Keogh.
\newblock The great time series classification bake off: a review and
  experimental evaluation of recent algorithmic advances.
\newblock {\em Data Mining and Knowledge Discovery}, 31(3):606--660, 2017.

\bibitem{bagnall15cote}
A.~Bagnall, J.~Lines, J.~Hills, and A.~Bostrom.
\newblock Time-series classification with {COTE}: The collective of
  transformation-based ensembles.
\newblock {\em {IEEE} Transactions on Knowledge and Data Engineering},
  27:2522--2535, 2015.

\bibitem{bostrom17binary}
A.~Bostrom and A.~Bagnall.
\newblock Binary shapelet transform for multiclass time series classification.
\newblock {\em Transactions on Large-Scale Data and Knowledge Centered
  Systems}, 32:24--46, 2017.

\bibitem{dau19ucr}
H.~Dau, A.~Bagnall, K.~Kamgar, M.~Yeh, Y.~Zhu, S.~Gharghabi, C.~Ratanamahatana,
  A.~Chotirat, and E.~Keogh.
\newblock The ucr time series archive.
\newblock {\em IEEE/CAA Journal of Automatica Sinica}, 6(6):1293--1305, 2019.

\bibitem{dempster19rocket}
Angus Dempster, Fran{\c{c}}ois Petitjean, and Geoffrey~I Webb.
\newblock Rocket: Exceptionally fast and accurate time series classification
  using random convolutional kernels.
\newblock {\em arXiv preprint arXiv:1910.13051}, 2019.

\bibitem{deng13forest}
H.~Deng, G.~Runger, E.~Tuv, and M.~Vladimir.
\newblock A time series forest for classification and feature extraction.
\newblock {\em Information Sciences}, 239:142--153, 2013.

\bibitem{fawaz19inception}
HI. Fawaz, B.~Lucas, G.~Forestier, C.~Pelletier, D.~Schmidt, J.~Weber, G.~Webb,
  L.~Idoumghar, P.~Muller, and F.~Petitjean.
\newblock Inceptiontime: Finding alexnet for time series classification.
\newblock {\em ArXiv}, 2019.

\bibitem{flynn19contract}
M.~Flynn, J.~Large, and A.~Bagnall.
\newblock The contract random interval spectral ensemble {(c-RISE)}: The effect
  of contracting a classifier on accuracy.
\newblock In {\em International Conference on Hybrid Artificial Intelligence
  Systems}, volume 11734 of {\em Lecture Notes in Computer Science}, pages
  381--392. 2019.

\bibitem{hills14shapelet}
J.~Hills, J.~Lines, E.~Baranauskas, J.~Mapp, and A.~Bagnall.
\newblock Classification of time series by shapelet transformation.
\newblock {\em Data Mining and Knowledge Discovery}, 28(4):851--881, 2014.

\bibitem{large19cawpe}
J.~Large, J.~Lines, and A.~Bagnall.
\newblock A probabilistic classifier ensemble weighting scheme based on cross
  validated accuracy estimates.
\newblock {\em Data Mining and Knowledge Discovery}, 33(6):1674--–1709, 2019.

\bibitem{lines15elastic}
J.~Lines and A.~Bagnall.
\newblock Time series classification with ensembles of elastic distance
  measures.
\newblock {\em Data Mining and Knowledge Discovery}, 29:565--592, 2015.

\bibitem{lines16hive}
J.~Lines, S.~Taylor, and A.~Bagnall.
\newblock {HIVE-COTE}: The hierarchical vote collective of transformation-based
  ensembles for time series classification.
\newblock In {\em Proc. 16th {IEEE} International Conference on Data Mining},
  2016.

\bibitem{lines18hive}
J.~Lines, S.~Taylor, and A.~Bagnall.
\newblock Time series classification with {HIVE-COTE}: The hierarchical vote
  collective of transformation-based ensembles.
\newblock {\em ACM Trans. Knowledge Discovery from Data}, 12(5), 2018.

\bibitem{middlehurst19scalable}
M.~Middlehurst, W.~Vickers, and A.~Bagnall.
\newblock Scalable dictionary classifiers for time series classification.
\newblock In {\em International Conference on Intelligent Data Engineering and
  Automated Learning}, Lecture Notes in Computer Science, pages 11--19. 2019.

\bibitem{rodriguez06rotf}
J.~Rodriguez, L.~Kuncheva, and C.~Alonso.
\newblock Rotation forest: A new classifier ensemble method.
\newblock {\em {IEEE} Transactions on Pattern Analysis and Machine
  Intelligence}, 28(10):1619--1630, 2006.

\bibitem{schafer15boss}
P.~Sch{\"a}fer.
\newblock The {BOSS} is concerned with time series classification in the
  presence of noise.
\newblock {\em Data Mining and Knowledge Discovery}, 29(6):1505--1530, 2015.

\bibitem{schafer12sfa}
P.~Sch{\"a}fer and M.~H{\"o}gqvist.
\newblock {SFA: a symbolic {Fourier} approximation and index for similarity
  search in high dimensional datasets}.
\newblock In {\em Proceedings of the 15th International Conference on Extending
  Database Technology}, pages 516--527, 2012.

\bibitem{shifaz19ts-chief}
A.~Shifaz, C.~Pelletier, F.~Petitjean, and G.~Webb.
\newblock {TS-CHIEF}: A scalable and accurate forest algorithm for time series
  classification.
\newblock {\em ArXiv e-prints}, arXiv:1906.10329, 2019.

\end{thebibliography}

\end{document}